%% file: main.tex
\documentclass{llncs}
\usepackage{makeidx}
\usepackage{graphicx}
\usepackage{caption}
\usepackage{array}
\usepackage{subfig}
\usepackage{color}
\usepackage[final]{changes}
\usepackage[bottom]{footmisc}

\definechangesauthor[name=Gullal Cheema, color=red]{GC}
\definechangesauthor[name=Saket Anand, color=blue]{SA}

\newcolumntype{P}[1]{>{\centering\arraybackslash}p{#1}}
\newcommand{\etal}{\emph{et al. }}

\begin{document}
	
\title{Automatic Detection and Recognition of Individuals in Patterned Species}

\author{Gullal Singh Cheema \and Saket Anand}

\institute{IIIT-Delhi, New Delhi, India \\ \email{\{gullal1408, anands\}@iiitd.ac.in}}

\maketitle

\begin{abstract}
Visual animal biometrics is rapidly gaining popularity as it enables a non-invasive and cost-effective approach for wildlife monitoring applications. Widespread usage of camera traps has led to large volumes of collected images, making manual processing of visual content hard to manage. In this work, we develop a framework for automatic detection and recognition of individuals in different patterned species like tigers, zebras and jaguars. Most existing systems primarily rely on manual input for localizing the animal, which does not scale well to large datasets.
In order to automate the detection process while retaining robustness to blur, partial occlusion, illumination and pose variations, we use the recently proposed Faster-RCNN object detection framework to efficiently detect animals in images. We further extract features from AlexNet of the animal's flank and train a logistic regression (or Linear SVM) classifier to recognize the individuals. We primarily test and evaluate our framework on a camera trap tiger image dataset that contains images that vary in overall image quality, animal pose, scale and lighting. We also evaluate our recognition system on zebra and jaguar images to show generalization to other patterned species. Our framework gives perfect detection results in camera trapped tiger images and a similar or better individual recognition performance when compared with state-of-the-art recognition techniques.
\end{abstract}

\input{sections/intro.tex}
\input{sections/rel_work.tex}

\input{sections/method.tex}
\input{sections/exp_res}
\input{sections/conclusion.tex}

{
\bibliographystyle{splncs03}
\bibliography{main}
}
\end{document}

%% file: sections/intro.tex
\section{Introduction}
Over the past two decades, advances in visual pattern recognition have led to many efficient visual biometric systems for identifying human individuals through various modalities like iris images \cite{daugman2004iris,tisse2002person}, facial images \cite{turk1991face,ahonen2006face} and fingerprints \cite{jain2000filterbank,jiang2000fingerprint}. Since the identification process relies on visual pattern matching, it is convenient and minimally invasive, which in turn makes it amenable to use with non-cooperative subjects as well. Consequently, visual biometrics has been applied to wild animals, where non-invasive techniques provide a huge advantage in terms of cost, safety and convenience. Apart from identifying or recognizing an individual, visual pattern matching is also used to classify species, detect occurrence or variation in behavior and also morphological traits. 

Historically, since mid-1900s, ecologists and evolutionary researchers have used sketch collections \cite{scott197817} and photographic records \cite{mizroch2003test,klingel1974zebra} to study, document and index animal appearance \cite{prodger2009darwin}. This is due to the fact that a large variety of animal species carry unique coat patterns like stripes on zebras and spots on jaguar. Even though the earlier studies provided ways to formalize unique animal appearance, manually identifying individuals is tedious and requires a human expert with specific skills, making the identification process prone to subjective bias. Moreover, as the volume of images increase, manual processing becomes prohibitively expensive. 

\begin{figure}[htp]
	\centering
	\includegraphics[width=.25\textwidth]{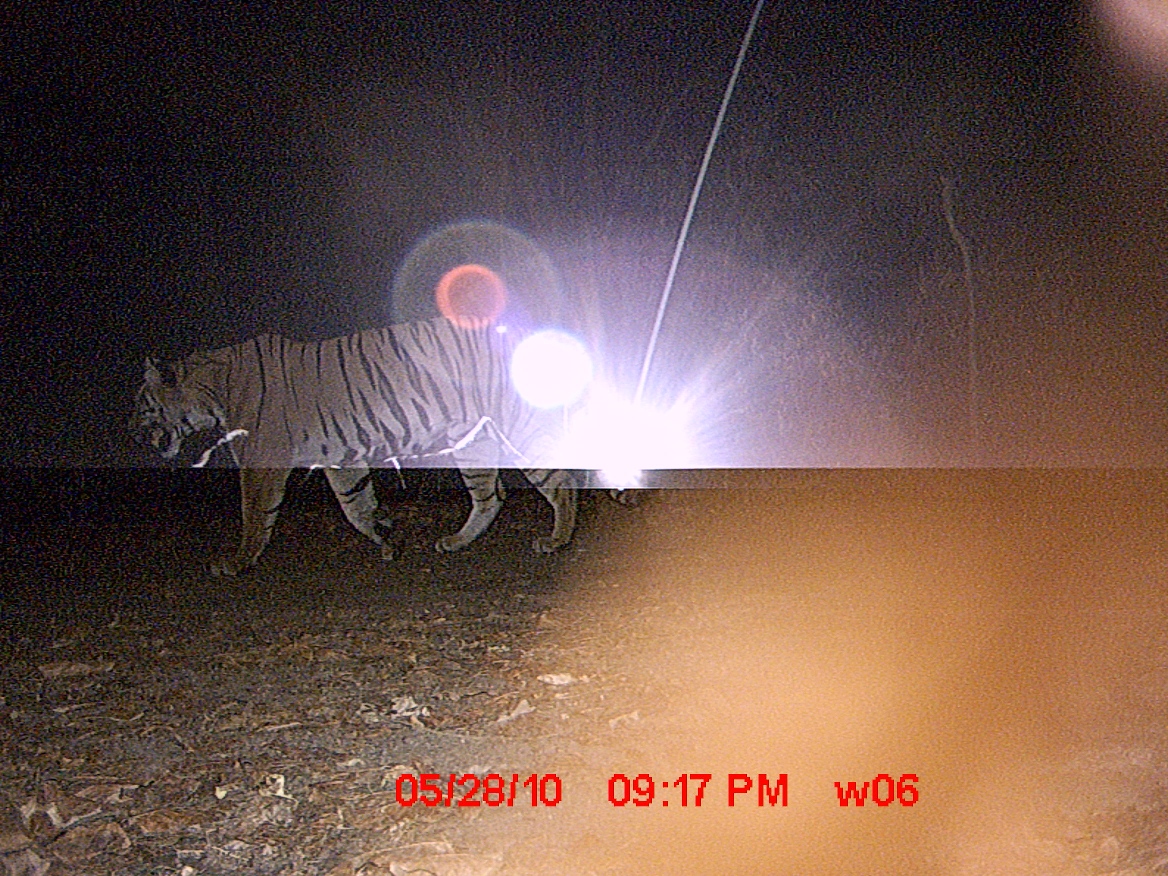}\quad
	\includegraphics[width=.25\textwidth]{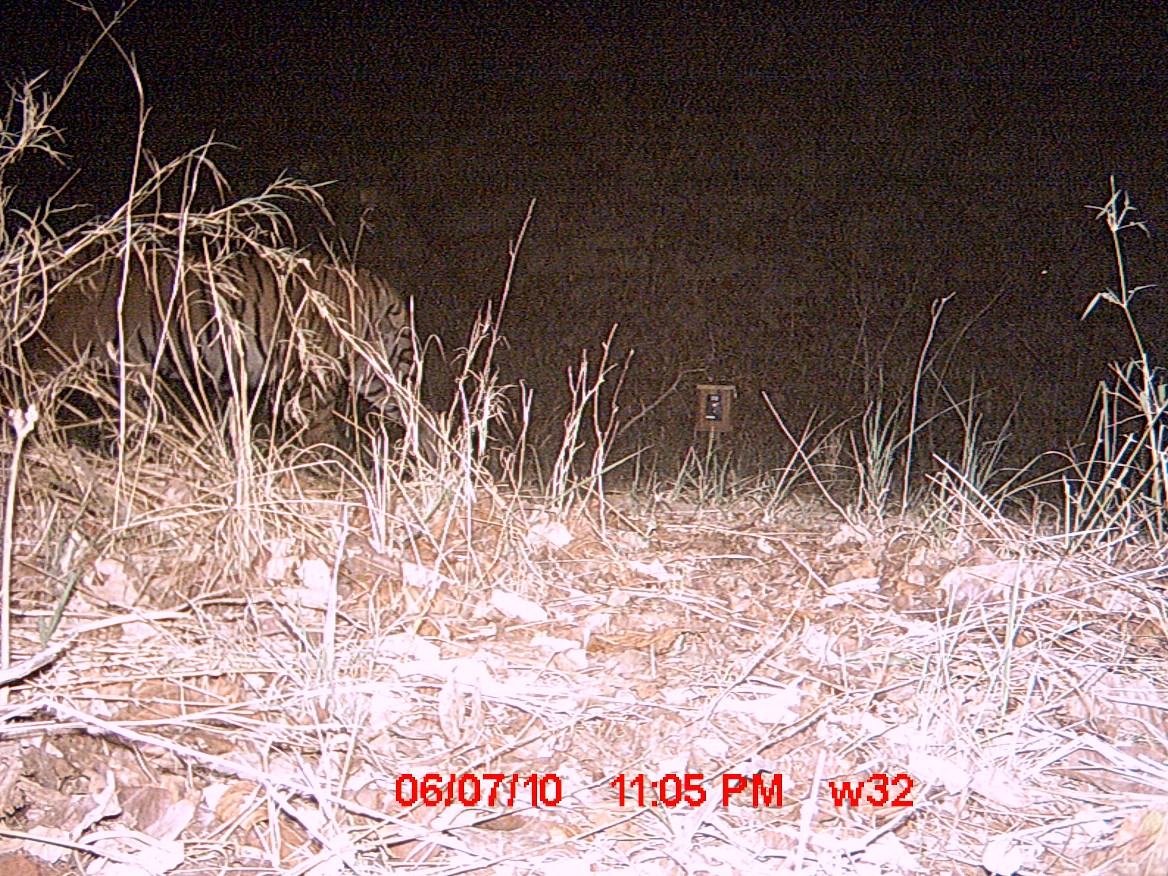}\quad
	\includegraphics[width=.25\textwidth]{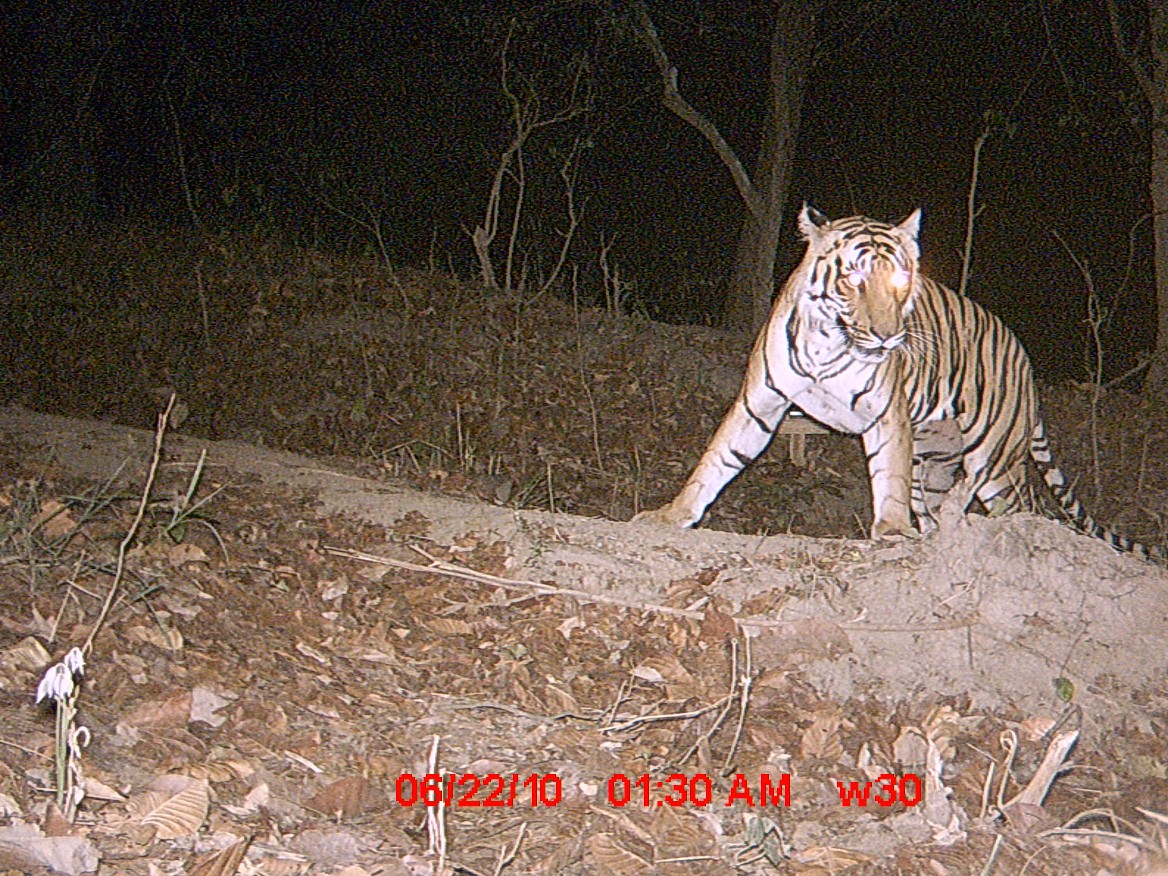}
	
	\medskip
	
	\includegraphics[width=.25\textwidth]{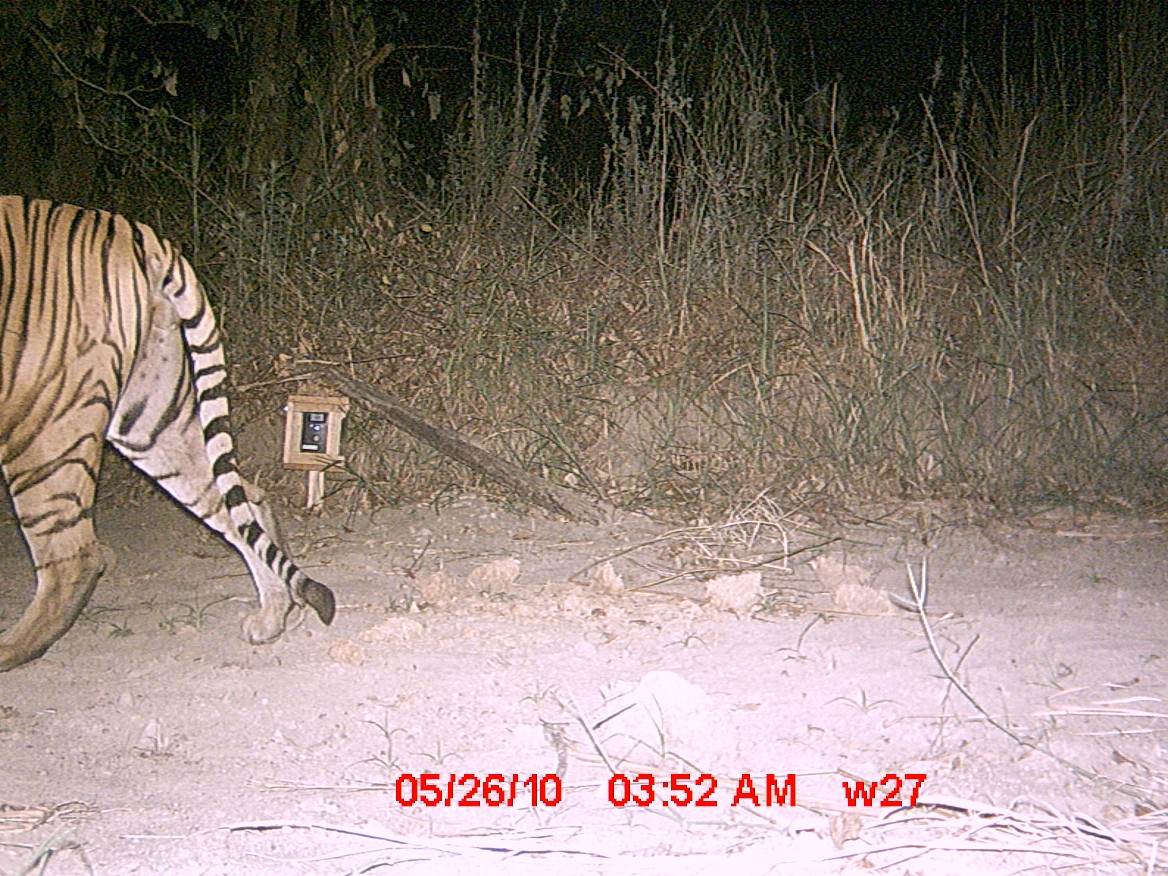}\quad
	\includegraphics[width=.25\textwidth]{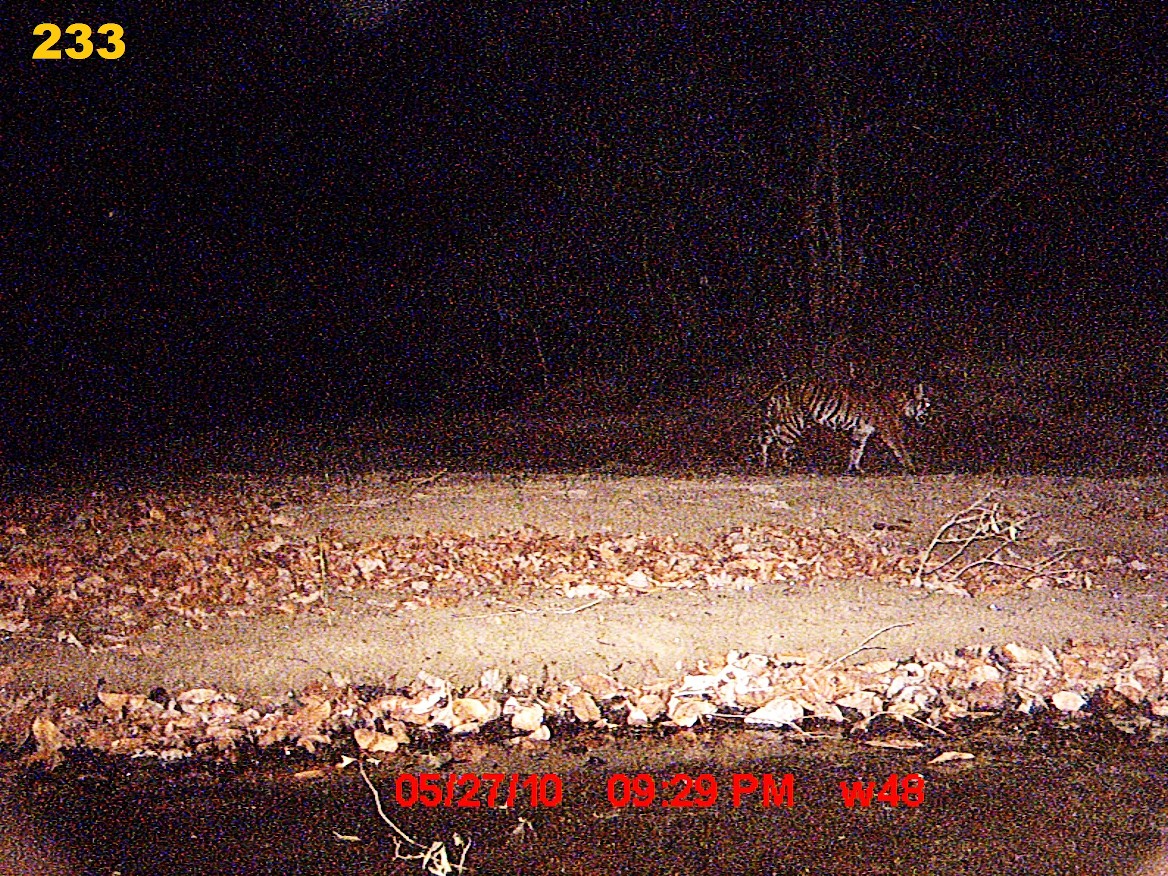}\quad
	\includegraphics[width=.25\textwidth]{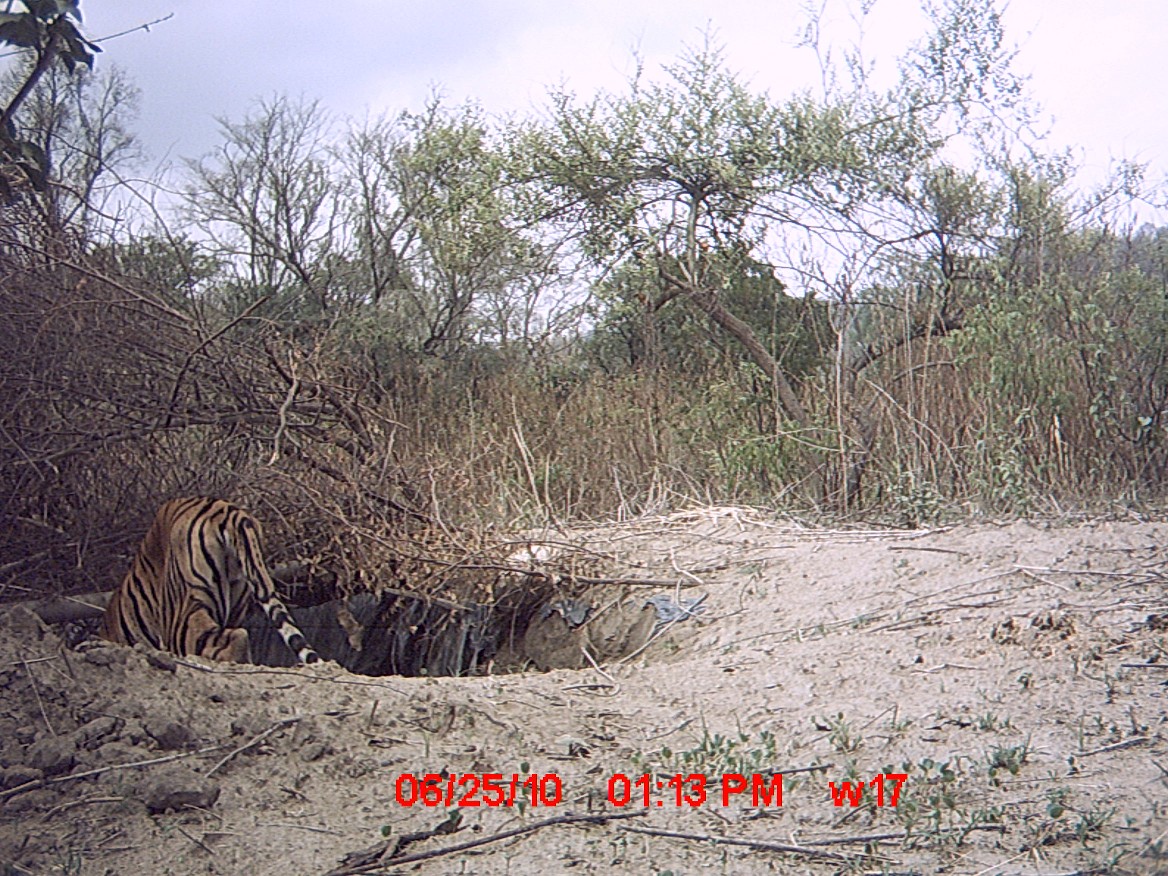}
	\caption{Sample challenging camera trapped images of tiger}
	\label{fig:samples1}
\end{figure}

With the advancement of computer vision techniques like object detection and localization \cite{felzenszwalb2010object}, pose estimation \cite{zhu2012face} and facial expression recognition \cite{cohen2003facial}, ecologists and researchers have an opportunity to systematically apply visual pattern matching techniques to automate wildlife monitoring. As opposed to traditional approaches like in field manual monitoring, radio collaring and GPS tracking, these approaches minimize the subjective bias, are repeatable, cost-effective, safer and less stressful for the human as well as the animal. However, unlike in the human case, there is little control over environmental factors during data acquisition of wild animals. Specifically, in the case of land animals, most of the image data is collected using camera traps fixed at probable locations where the animal of interest can be located. Due to these reasons, the recognition systems have to be robust enough to work on images with drastic illumination changes, blurring and occlusion due to vegetation. For example, some of the challenging images in our tiger dataset can be seen in the Fig. \ref{fig:samples1}.

In recent years, organizations like WWF-India and projects like \textit{Snapshot Serengeti} project \cite{swanson2015snapshot} have gathered and cataloged millions of images through hundreds of camera trap sites spanning large geographical areas.  With this unprecedented increase in quantity of camera trap images, there is a requirement of visual monitoring systems that can automatically sort and organize images based on a desired category (species/individual level) in short amount of time. Also, many of the animal species are endangered and require continuous monitoring, especially in areas where they are vulnerable to poaching, predators and already less in number. Such monitoring efforts can help to protect animals, maintain population across different geographical areas and also protect the local ecosystem. 

In this work, we develop a framework for detecting and recognizing individual patterned species that have unique coat patterns such as stripes on zebras, tigers and spots on Jaguars. State-of-the-art systems such as Extract-Compare \cite{hiby2009tiger} by Hiby \etal and HotSpotter \cite{crall2013hotspotter} by Crall \etal work well, but require user input for every image and hence fail to scale to large datasets. Automatic detection methods proposed in \cite{burghardt2006real,burghardt2004tracking} detect smaller patches on animals but not the complete animal, and are sensitive to lighting conditions and multiple instances of animals in the same image. In this work, we use the recently proposed convolutional neural network (CNN) based detector, Faster-RCNN \cite{ren2015faster} by Ren \etal that is able to detect different objects at multiple scales. The advantage of using a deep CNN based architecture is robustness to illumination and pose variations as well as location invariance, which proves to be very effective for localizing animals in images in uncontrolled environments. We use Faster-RCNN to detect the body and flank region of the animal and pass it through a pre-trained AlexNet \cite{krizhevsky2012imagenet} to extract discriminatory features, which is used by a logistic regression classifier for individual recognition.

The remainder of the paper is structured as follows. In Section \ref{sec:rel_work}, we will briefly talk about recent related work in animal detection and individual animal recognition. Section \ref{sec:bg} lays the groundwork for the description of our proposed framework in Section \ref{sec:method}. We then present empirical results on data sets of various patterned species, and report performance comparisons in Section \ref{sec:expt} before concluding in Section \ref{sec:conc}.

%% file: sections/rel_work.tex
\section{Related Work}\label{sec:rel_work}
In this section we briefly discuss recent advances in animal species and individual identification, focusing on land animals exhibiting unique coat patterns.

\subsection{Animal Detection}
One of the earliest works on automatic animal detection \cite{burghardt2006real,burghardt2004tracking} uses Haar-like features and a low-level feature tracker to detect a lion's face and extract information to predict its activity like still, walking or trotting. The system works in real time and is able to detect faces at multiple scales although only with \emph{slight} pose variations. Zhang \etal \cite{zhang2011tiger} detect heads of animals like tiger, cat, dog, cheetah, etc. by using shape and texture features to improve image retrieval. The approach relies on prominent `pointed' ear shapes in frontal poses which makes it sensitive to head-pose variations. These approaches rely on identifying different parts of the animal to detect and track an individual, but are likely to fail in case of occlusion or significant pose change. 

CNNs are known to be robust to occlusion and pose variations and have also been used to automatically learn discriminatory features from the data to localize Chimpanzee faces \cite{freytag2016}. Also, recently Norouzzadeh \etal \cite{norouzzadeh2017automatically} used various CNN architectures like Alexnet \cite{krizhevsky2012imagenet}, VGGnet \cite{simonyan2014very} and ResNet \cite{he2016deep} to classify 48 animal species using the Snapshot Serengeti \cite{swanson2015snapshot} dataset with 3.2 million camera trap images and achieved $\sim$96\% classification accuracy.

\subsection{Individual Animal Recognition}
Hiby \etal \cite{hiby2009tiger} developed `Extract-Compare', one of the first interactive software tools for recognizing individuals by matching coat patterns of species like tiger, cheetah, giraffe, frogs, etc. The tool works in a retrieval framework, where a user inputs a query image and individuals with similar coat patterns are retrieved from a database for final verification by the user. Prior to the pattern matching, a coarsely parametric 3D surface model is fit on the animal's body, e.g., around the flank of a tiger, or the head of an armadillo. This surface model fitting makes the pattern matching robust to animal and camera pose. However, in order to fit the 3D surface model, the user has to carefully mark several key points like the head, tail, elbows, knees, etc. While this approach works well in terms of accuracy, it is not scalable to a large number of images as the manual processing time for an image could be as high as 30 seconds.  

Lahiri \etal introduced StripeSpotter \cite{lahiri2011biometric} that extracts features from flanks of a Zebra as 2D arrays of binary values. This 2D array depicts the white and black stripe pattern which can be used to uniquely identify a zebra. The algorithm uses a dynamic programming approach to calculate a value similar to edit distance between two strings. Again, the flank region is extracted manually and each query image is matched against every other image in the database.

HotSpotter \cite{crall2013hotspotter} and Wild-ID \cite{bolger2012wildid} use SIFT \cite{lowe2004distinctive} features to match query image of the animal with a database of existing animals. Both the tools require a manual input for selecting the region of interest so that SIFT features are unaffected by background clutter in the image. In addition to matching each query image descriptor against each database image separately, Hotspotter also uses \textit{one vs many} approach by matching each query image descriptor with all the database descriptors. It uses efficient data structure such as a forest of kd-trees and different scoring criterion to efficiently find the approximate nearest neighbor. Hotspotter also performs spatial re-ranking to filter out any spatially inconsistent descriptor matches by using RANSAC solution used in \cite{philbin2007object}. However, spatial re-ranking does not perform better than simple \textit{one vs many} matching.

%% file: sections/method.tex
\section{Background}\label{sec:bg} 
In this section we briefly describe the deep neural network architectures that we employ in our animal detection and individual recognition framework.

\subsection{Faster-RCNN}
Faster-RCNN \cite{ren2015faster} by Ren \etal is a recently proposed object detection technique that is composed of two modules in a single unified network. The first module is a deep CNN that works as a Region Proposal Network (RPN) and proposes regions of interest (ROI), while the second module is a Fast R-CNN \cite{girshick2015fast} detector that categorizes each of the proposed ROIs. This unification of RPN with the detector lowers test-time computation without noticeable loss in detection performance. 

The RPN takes input an image of any size and outputs a set of rectangular object proposals, each with an objectness (object vs background) score. In addition to shared convolutional layers, RPN has an additional small network with one $n$ x $n$ convolutional layer and two sibling fully connected layers (one for box-regression and one for box-classification). At each sliding-window location for the $n$ x $n$ convolution layer, multiple region proposals (called anchors) are predicted with varying scales and aspect ratios. Each output is then mapped to lower-dimensional feature which is then fed into the two sibling layers. 

The Fast R-CNN detection network on the other hand can be a ZF \cite{zeiler2014visualizing} or VGG \cite{simonyan2014very} net which, in addition to shared convolution layers, has two fully connected (fc6 and fc7) layers and two sibling class score and bounding box prediction fully connected layers. For further details on cost functions and training of Faster-RCNN, see \cite{ren2015faster}. We discuss training, hyperparameter setting and implementation details specific to tiger detection in Sec. \ref{sec:method} and Sec. \ref{sec:expt}.

\subsection{AlexNet}
AlexNet was proposed in \cite{krizhevsky2012imagenet} with five convolutional and three fully-connected layers. With 60 million parameters, the network was trained using a subset of about 1.2 million images from the ImageNet dataset for classifying about 150,000 images into 1000 different categories. The success of AlexNet on a large-scale image classification problem led to several works that used \emph{pre-trained} networks for feature representations which are fed to an application specific classifier. We follow a similar approach for recognition of individuals in patterned species, with a modification of the input size and consequently the feature map dimensions.

\section{Methodology}\label{sec:method} 
\begin{figure}[htp]
	\centering
	\includegraphics[width=1\textwidth]{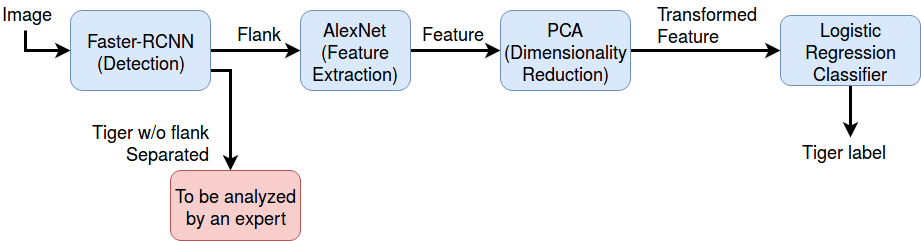}
	\caption{Proposed framework for animal detection and individual recognition}
	\label{fig:framework}
\end{figure}

In this work we address two problems in animal monitoring: First is to detect and localize the patterned species in a camera trap image, and the second is to uniquely identify the detected animal against an existing database of the same species. The proposed framework can be seen in the Fig. \ref{fig:framework}.

\subsection{Data Augmentation} 
To increase the number of images for training phase and avoid over-fitting, we augment the given training data for both detection and individual recognition. For detection, we double the number of training images by horizontally flipping (mirroring) each image while training the Faster-RCNN. 

In case of recognizing individuals, the number of training samples is very small because of relatively few side pose captures per tiger. \deleted[id=GC,remark=This is the only text I found which could be skipped]{In a typical data acquisition process performed annually, camera traps are installed over a local forest area of several hundreds of square kilometers and remain operational for 40-60 days \cite{jhala2015}. In this duration, an individual may appear only on a few days, typically yielding less than a few tens of images per individual.} 
Therefore, in order to learn to classify individual animals, we need stronger data augmentation techniques. We use contrast enhancement and random filtering (Gaussian or median) for each training image increasing our training set to thrice the number of training images originally.
 
\subsection{Detection using Faster-RCNN} 
We detect both the tiger and the flank region using Faster-RCNN. During training, both the image and the bounding boxes (tiger and flank) are input to the network. The bounding boxes for the flanks are given for only those images in which the flank is not occluded and distorted due to the pose of the tiger. The network is trained to detect 3 classes: tiger, flank and the background. \deleted[id=GC, remark=All parameters as used in the original Faster-RCNN paper]{All the images are re-scaled to 1000 x 600 as in the original implementation. For region proposals (anchors), we use 3 scales with box areas of $128^2$, $256^2$ and $512^2$ pixels, and 3 aspect ratios of 1:1, 1:2, and 2:1 as used originally in \cite{ren2015faster}.} All the parameters used for training are as used in the original implementation.

For training, the whole network is trained with 4-step alternating training mentioned in \cite{ren2015faster}. We use ZF \cite{zeiler2014visualizing} net in our framework which has five shareable convolutional layers. In the first step, RPN is trained end-to-end for the region proposals task by initializing the network with an ImageNet-pre-trained model. Fast R-CNN is then trained in the second step with weights initialized by ImageNet-pre-trained model and by using the proposals generated by step-1 RPN. Weight sharing is performed in the third and fourth step, where RPN training is initialized with detector network and by fixing the shared convolutional layers, only layers unique to RPN are fine-tuned. Similarly, Fast R-CNN is trained in the fourth step by fixing the shared layers and fine-tuning only the unique layers of the detector. Additionally, we also fix first two convolutional layers in first two steps of the training for tiger detection as the initial layers are already fine-tuned to detect low-level features like edges. 

During testing, only an image is input to the network and it outputs the bounding boxes and the corresponding objectness scores. As Faster-RCNN outputs multiple bounding boxes per category, some of which are highly overlapping, non-maximum suppression (NMS) is applied to reduce the redundant boxes. Because the convolution layers are shared, we can test the image in one go in very less time (0.3-0.6 seconds/image on a GPU).

\subsection{Identification}
For identification, we only use flank regions because they contain the discriminatory information to uniquely identify the patterned animals. The images in which the tiger is detected, but the flank is not, are separated to be analyzed by the expert. A tool such as Extract-Compare \cite{hiby2009tiger} can be used for difficult cases with extreme pose or occlusion.

We use ImageNet-pre-trained AlexNet \cite{krizhevsky2012imagenet} to extract features from the flank region and train a logistic regression classifier to recognize the individuals. While this deviates from the end-to-end framework, typical of deep networks, our choice of this approach was to resolve the problem of very low training data for identifying individuals. We tried fine-tuning AlexNet with our data, however, the model overfitted the training set. For feature representation, we used different convolutional layers and fully connected layers to train our classifier and obtained the best results with the third convolutional layer (conv3). Since ImageNet is a large-scale dataset, the pre-trained weights of AlexNet in the higher layers are not optimized for a fine-grained task such as individual animal recognition. On the other hand, the middle layers (like conv3) capture interactions between edges and are discriminative enough to give good results for our problem. 

To minimize distortion introduced by resizing the detected flank region to a unit aspect ratio of AlexNet ($ 227 \times 227 $), we modify the size of input to AlexNet and hence the subsequent feature maps. Since the conv3 feature maps are high dimensional, we apply a PCA (Principal Component Analysis) based dimensionality reduction and use principal components that explain 99\% of the energy.

%% file: sections/exp_res.tex
\section{Experiments}\label{sec:expt}
All experiments are carried out with Python (and PyCaffe) running on i7-4720HQ 3.6GHz processor and Nvidia GTX-950M GPU. For Faster-RCNN \cite{ren2015faster} training, we use a server with Nvidia GTX-980 GPU. We used the python implementation of Faster-RCNN \footnote{\url{https://github.com/rbgirshick/py-faster-rcnn}} and labelImg \footnote{\url{https://github.com/tzutalin/labelImg}} annotation tool for annotating the tiger and jaguar images. We also use python's sklearn library for logistic regression classifier. Over three different datasets, we compare our results with HotSpotter \cite{crall2013hotspotter}, which showed superior performance as compared to Wild-ID \cite{bolger2012wildid} and StripeSpotter \cite{lahiri2011biometric}.

\subsection{Datasets}
\noindent\textbf{Tiger Dataset}: The dataset is provided by Wildlife Institute of India (WII) and contains about 770 images captured from camera traps. The images as shown in Fig. \ref{fig:samples1} are very challenging due to severe viewpoint and illumination changes, motion blur and occlusions. We use this for both detection and individual recognition. \\
\noindent\textbf{Plains Zebra Dataset}\footnote{\url{http://compbio.cs.uic.edu/$\sim$stripespotter/}} was used in StripeSpotter \cite{lahiri2011biometric}. The stripe patterns are less discriminative than tigers, however, the images in this dataset have little viewpoint and appearance variations as most images were taken within seconds of each other. We use the cropped flank regions provided in the dataset for comparison with hotspotter.\\
\noindent\textbf{Jaguar Dataset}\footnote{Provided by Marcella J Kelly upon request: \url{http://www.mjkelly.info/}} is a smaller dataset also obtained from camera traps, but have poorer image quality (mostly night images), and moderate viewpoint variations.

We summarize the three datasets and our model parameters for the individual recognition task in Table \ref{tab:data}.
\begin{table}[!h]
	\centering
	\begin{tabular}{ |P{2cm}|P{1.5cm}|P{1.5cm}|P{2cm}|P{2cm}|P{1.5cm}|} 
		\hline
		\textbf{Species} & \textbf{\#Images} & \textbf{\#Labels} & \textbf{Feature size conv3} & \textbf{Feature size after PCA} & \textbf{C} \\
		\hline
		Tiger & 260 & 44 & 63360 & $\sim$180 & $1e6$ \\ 
		Plains Zebra & 821 & 83 & 40320 & $\sim$460 & $1e6$ \\
		Jaguar & 112 & 37  & 63360 & $\sim$70 & $1e5$ \\
		\hline
	\end{tabular}
	\caption{Dataset statistics and model parameters. \textbf{C} is the inverse of regularization strength used in logistic regression classifier.}
	\label{tab:data}
	\vspace{-1cm}
\end{table}
\begin{figure}[htp]
	\centering
	\includegraphics[width=.23\textwidth]{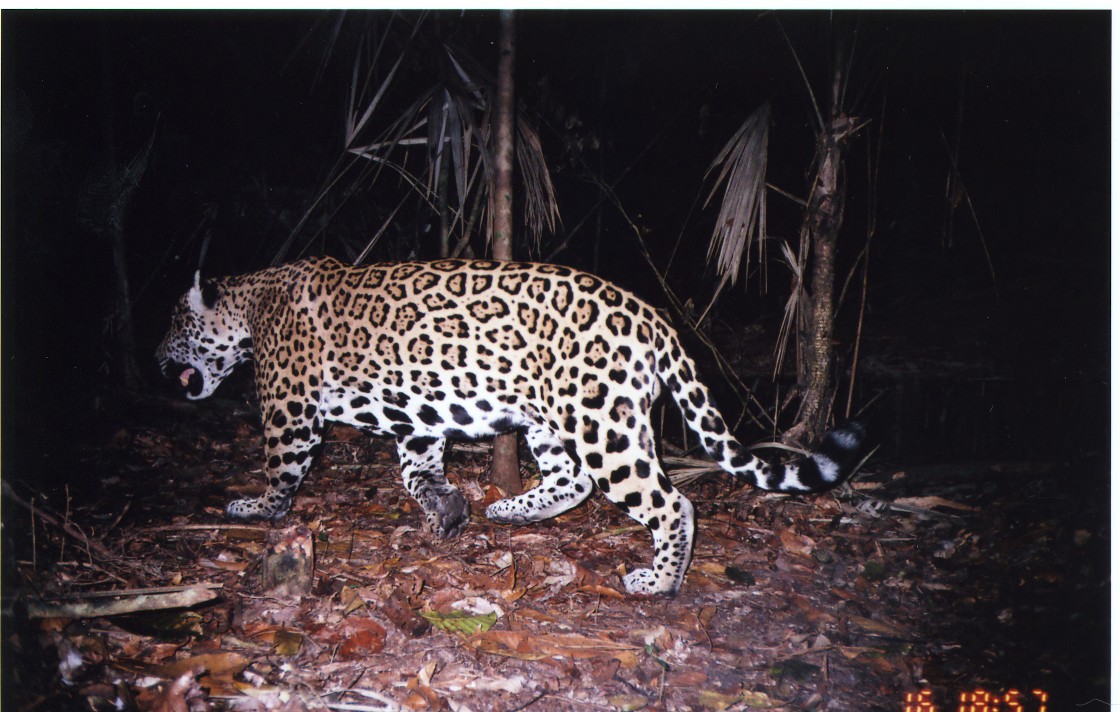}\quad
	\includegraphics[width=.23\textwidth]{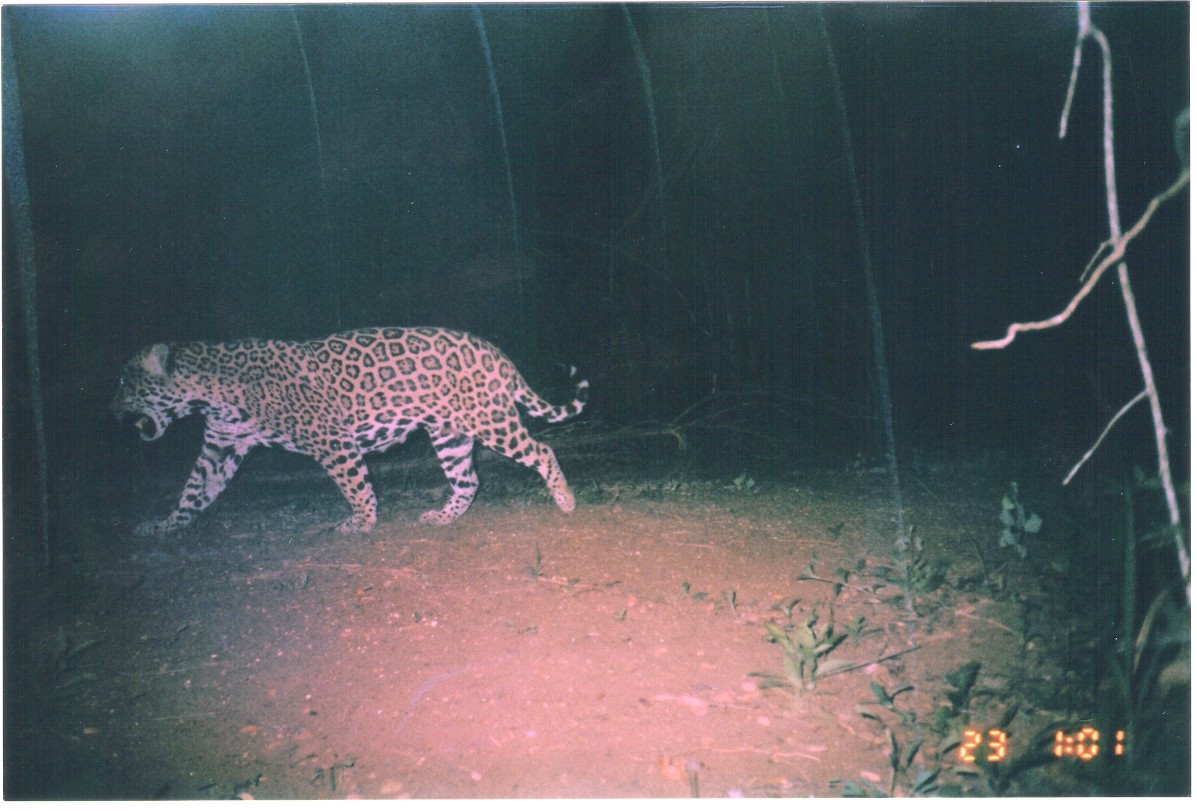}\quad
	\includegraphics[width=.23\textwidth]{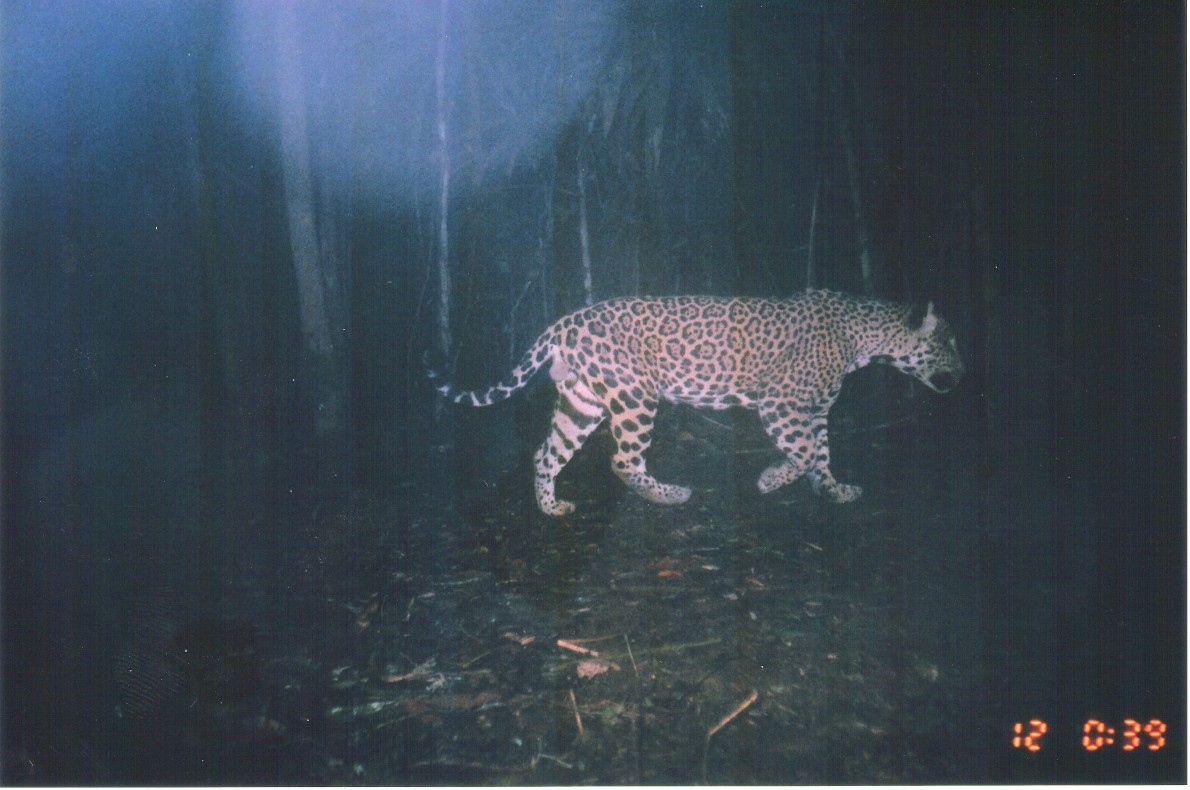}
	
	\medskip
	
	\includegraphics[width=.23\textwidth]{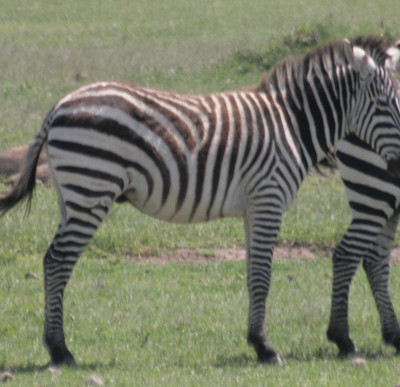}\quad
	\includegraphics[width=.23\textwidth]{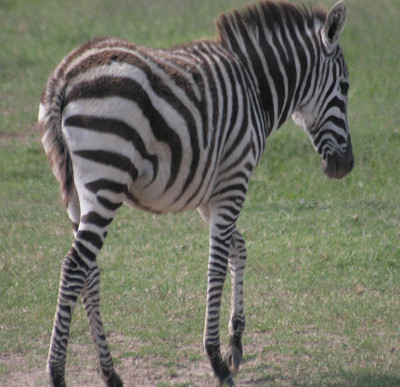}\quad
	\includegraphics[width=.23\textwidth]{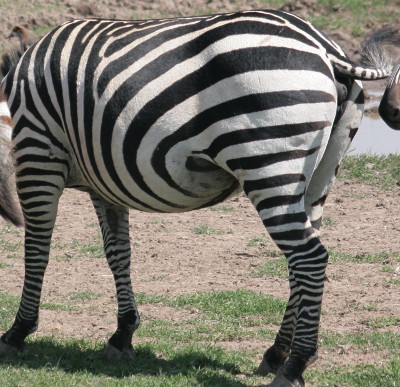}
	
	\caption{Sample images from the other two datasets. Row 1: Jaguars. Row 2: Plains Zebra.}
	\label{fig:samples2}
\end{figure}

\subsection{Detection}
We use 687 tiger images for training and testing the detection system after removing the ones in which the tiger is hardly visible (only tail) and a few very poor quality images (very high contrast due to flash/sun rays). We divide the data for training and testing with a split of 75\%/25\% respectively into a disjoint set of tigers. With data augmentation, we have a total of 1032 (516x2) images in the training set and 171 in the testing set.

For training Faster-RCNN, we randomly initialize all new layers by drawing weights from a zero-mean Gaussian distribution with standard deviation 0.01. We fine-tune RPN in both step 1 and 3 for 12000 iterations and Fast-RCNN in both step 2 and 4 for 10000 iterations. We use a learning rate of 0.001 for 10k and 8k mini-batches respectively, and 0.0001 for the next 2k mini-batches. We use a mini-batch size 1 (RPN) and 2 (Fast-RCNN) images,  momentum of 0.9 and a weight decay of 0.0005 as used in \cite{ren2015faster}. For applying non-maximum suppression (NMS), we fix the NMS threshold at 0.3 (best) on predicted boxes with objectness score more than 0.8, such that all the boxes with IoU greater than the threshold are suppressed.

\begin{figure}[!h]
	\centering
	\includegraphics[width=.35\textwidth]{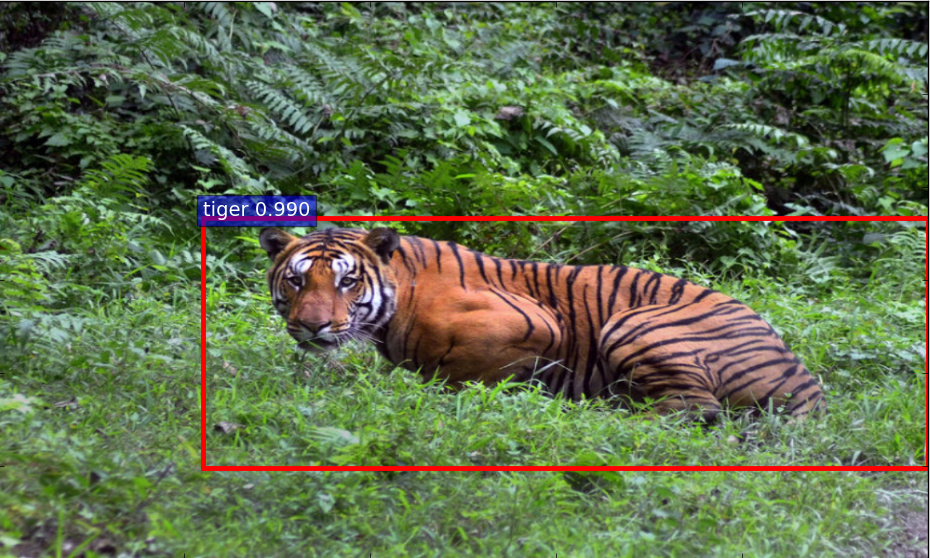}\quad
	\includegraphics[width=.35\textwidth]{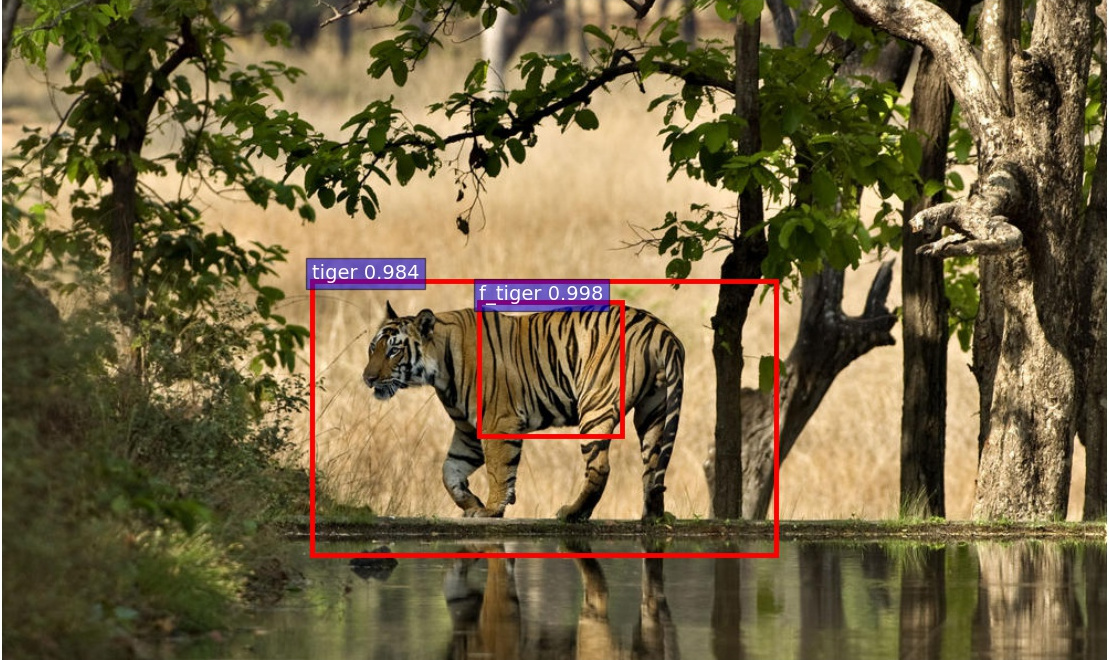}
	
	\medskip
	
	\includegraphics[width=.35\textwidth]{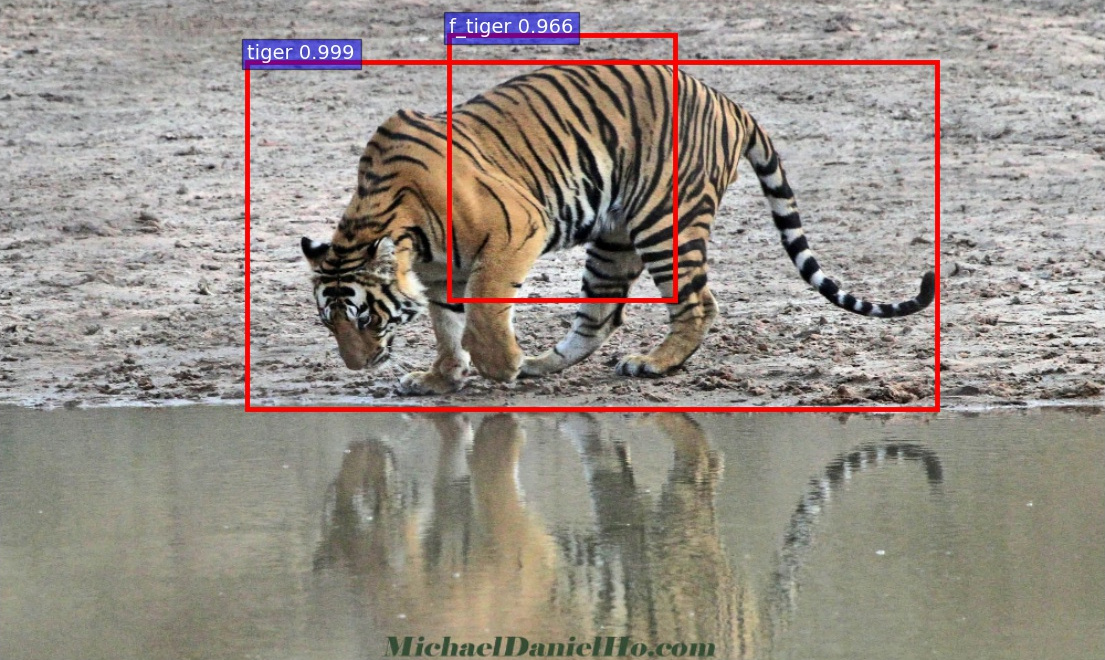}\quad
	\includegraphics[width=.35\textwidth]{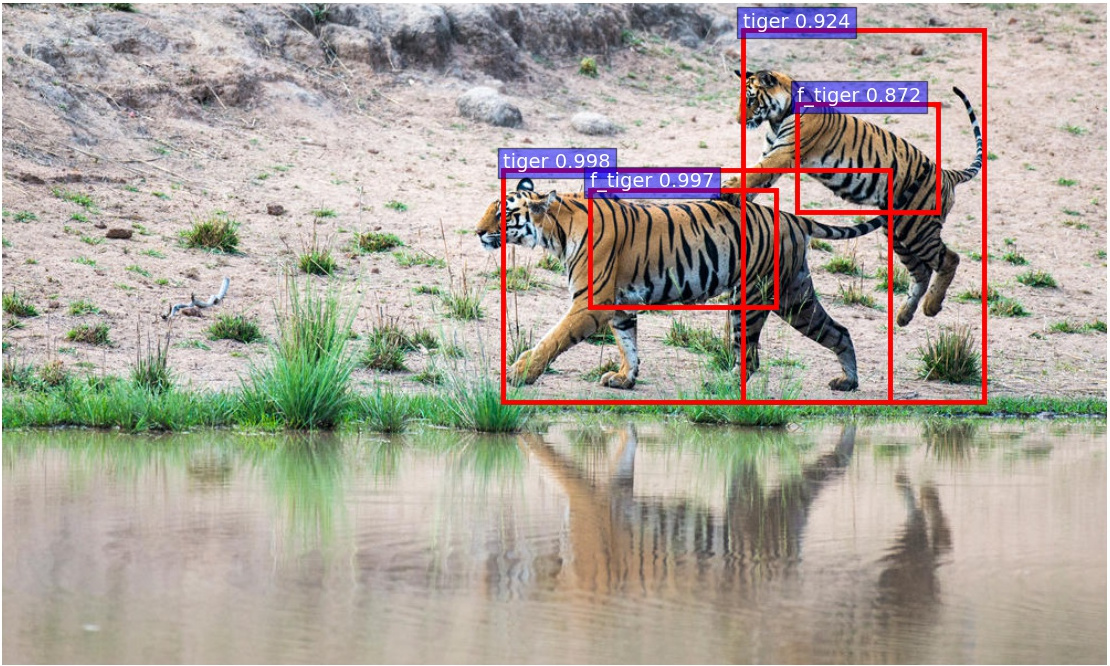}
	
	\caption{Qualitative detection results on images taken from the Internet. The detected boxes are labeled as (Label: Objectness score).}
	\label{fig:detres}
\end{figure}

We report Average Precision (AP) and mean AP for tiger and flank detection, which is a popular metric used for object detection. \deleted[id=GC, remark=This can be deleted as it is a popular metric]{It is calculated by first taking an average of (IoU) between the ground truth and predicted bounding box for all images and then taking mean over the number of categories.} 
The results for tiger and flank detection with varying NMS threshold are reported in Table \ref{tab:detres}. With increasing NMS threshold, number of output bounding boxes also increase which leads to poor detection results. 
\begin{table}[h]
	\centering
	\begin{tabular}{ |P{2.5cm}|P{1cm}|P{1cm}|P{1cm}|P{1cm}|P{1cm}|P{1cm}|P{1cm}|P{1cm}| } 
		\hline
		\textbf{Object/NMS threshold} & \textbf{0.2} & \textbf{0.3} & \textbf{0.4} & \textbf{0.5} & \textbf{0.6} & \textbf{0.7} & \textbf{0.8} & \textbf{0.9} \\
		\hline
		Tiger & 90.7 & \textbf{90.9} & 90.6 & 90.3 & 88.9 & 85.3 & 73.5 & 45.2 \\
		Flank & 90.6 & \textbf{90.6} & 90.4 & 89.4 & 87.2 & 76.9 & 57.0 & 41.9 \\ \hline \hline
		\textbf{mean AP} & 90.6 & \textbf{90.7} & 90.5 & 89.9 & 88.0 & 81.1 & 65.2 & 43.6 \\
		\hline
	\end{tabular}
	\caption{Results for tiger and flank detection}
	\label{tab:detres}
\end{table}
We also show some qualitative results on tiger images taken from the Internet, which are quite different in quality and background when compared to the camera trap images as shown in Fig. \ref{fig:detres}.
\subsection{Individual Recognition}
\begin{figure}[!h]
	\centering
	\subfloat[Zebra dataset]{\label{fig:z}\includegraphics[width=0.33\textwidth]{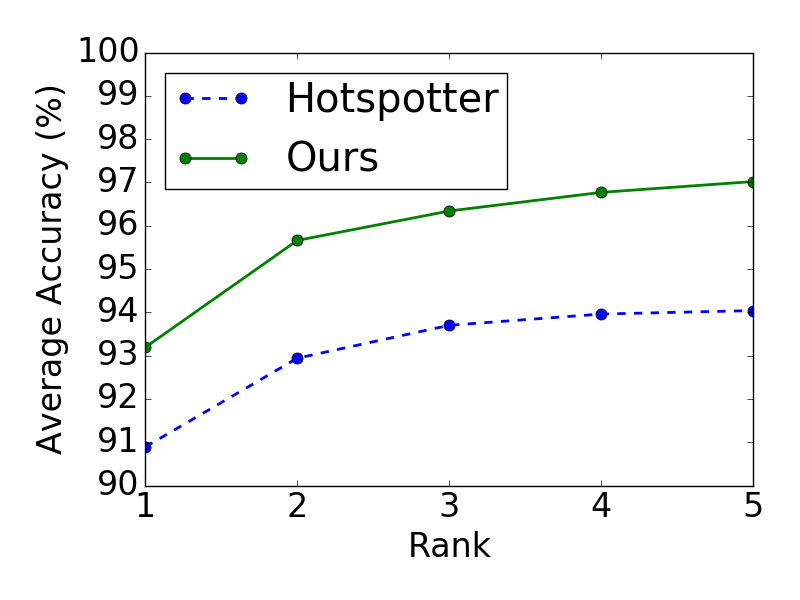}}
	\subfloat[Tiger dataset]{\label{fig:t}\includegraphics[width=0.33\textwidth]{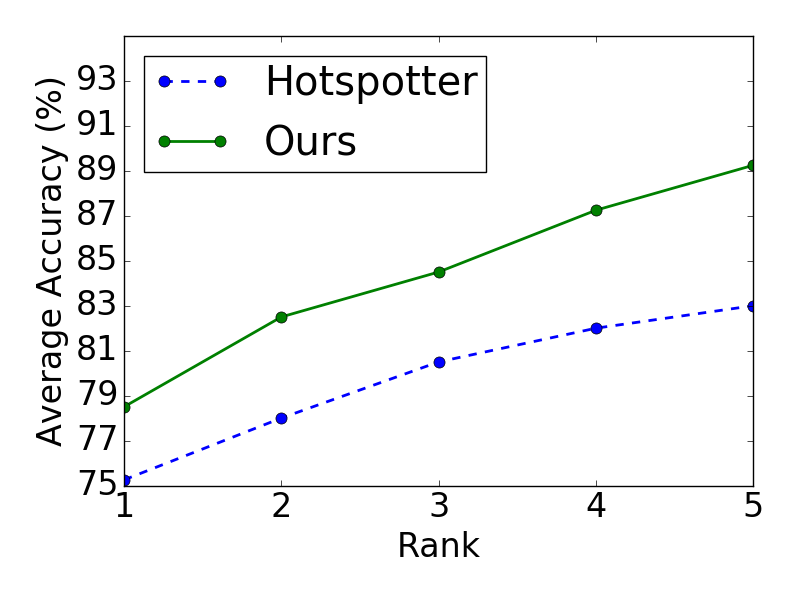}}
	\subfloat[Jaguar dataset]{\label{fig:j}\includegraphics[width=0.33\textwidth]{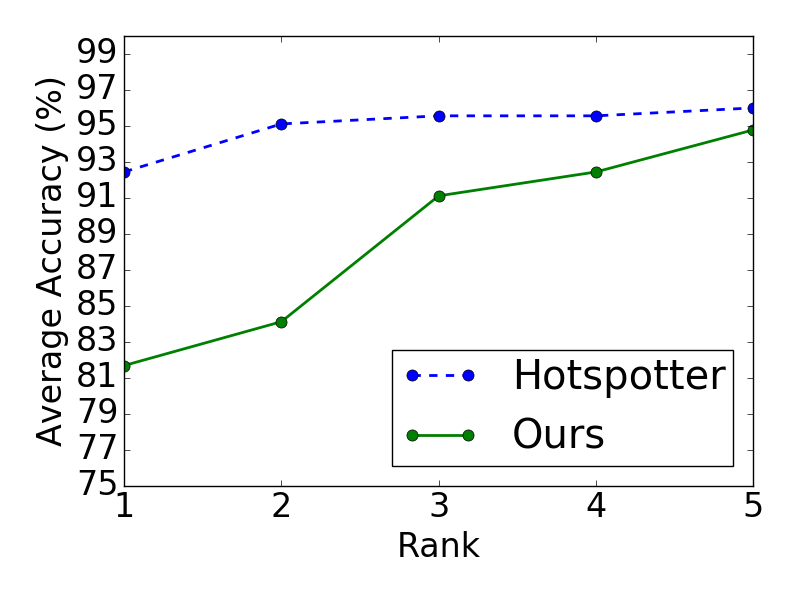}}
	\caption{CMC curve comparison}
	\label{fig:cmcs}
\end{figure}

We use conv3 features of AlexNet for training a logistic regression classifier to classify individuals. For each dataset, we generate five random splits with 75\% for training and 25\% for testing. For our framework, flanks of tiger and jaguar are resized to 256 x 192 and for the zebra to 256 x 128 which is equivalent to average size of flank images for the respective dataset. We learn a logistic regression model with $\ell_1$ regularization and perform grid search to find the parameter C. Specific data statistics and model parameters are reported in Table \ref{tab:data}. We compare our results with HotSpotter and report the average rank 1 accuracy for all the datasets in Table \ref{tab:idres}. In Fig. \ref{fig:cmcs}, we show the Cumulative Match Characteristic (CMC) curves from rank 1 to rank 5 for our method compared with Hotspotter over all the datasets.
\begin{table}[!h]
	\centering
	\begin{tabular}{ |P{2cm}|P{2cm}|P{2cm}|P{2cm}| } 
		\hline
		\textbf{Dataset} & \textbf{Ours (227x227)} & \textbf{Ours (Resized)} & \textbf{HotSpotter} \\
		\hline
		Tiger & 76.5 $\pm$ 2.2 & \textbf{80.5 $\pm$ 2.1} & 75.3 $\pm$ 1.2 \\
		Jaguar & 73.5 $\pm$ 1.8 & 78.6 $\pm$ 2.3 & \textbf{92.4 $\pm$ 1.1} \\
		Zebra & 91.1 $\pm$ 1.2 & \textbf{93.2 $\pm$ 1.4} & 90.9 $\pm$ 0.8 \\
		\hline
	\end{tabular}
	\caption{Average rank 1 accuracy comparison}
	\label{tab:idres}
	\vspace{-0.5cm}
\end{table}
The CMC curves indicate that the CNN based architecture clearly works better than HotSpotter in case of stripe patterns, even as we compare lower-rank accuracies. In the jaguar dataset, Hotspotter has a much higher rank-1 accuracy, but we observe a rising trend of our deep learning based approach as we compare lower-rank accuracies. We conjecture that the pre-trained AlexNet feature representation is not as discriminative for spots in jaguars as in case of stripes in tigers or zebras.

%% file: sections/conclusion.tex
\section{Conclusion}\label{sec:conc}
In this paper, we proposed a framework for automatic detection and individual recognition in patterned animal species. We used the state-of-the-art CNN based object detector Faster-RCNN \cite{ren2015faster} and fine-tuned it for the purpose of detecting the whole body and the flank of the tiger. We then used the detected flanks and extracted features from a pre-trained AlexNet \cite{krizhevsky2012imagenet} to train a logistic regression classifier for classifying individual tigers. We also performed individual recognition task on zebras and jaguars. We get perfect results for tiger detection and perform better than Hotspotter \cite{crall2013hotspotter} while comparing rank-1 accuracy for  individual recognition for tiger and zebra images. Even though AlexNet \cite{krizhevsky2012imagenet} features used for individual recognition are trained on Imagenet data, they seem to be as robust as SIFT \cite{lowe2004distinctive} features as shown by our quantitative results. We plan do a thorough comparison in future with larger datasets to obtain deeper insights. For jaguar images, Hotspotter works better at rank-1 accuracy, but the proposed method shows improving trends as we compare lower-rank accuracies. 

\section{Acknowledgments}
The authors would like to thank WII for providing the tiger data, Infosys Center for AI at IIIT-Delhi for computing resources and the anonymous reviewers for their invaluable comments.